\documentclass[sigconf]{acmart}
\AtBeginDocument{%
  \providecommand\BibTeX{{%
    \normalfont B\kern-0.5em{\scshape i\kern-0.25em b}\kern-0.8em\TeX}}}

\usepackage{fancyhdr}
\renewcommand\footnotetextcopyrightpermission[1]{}

\setcopyright{acmcopyright}
\copyrightyear{2018}
\acmYear{2018}
\acmDOI{XXXXXXX.XXXXXXX}

\acmConference[Conference acronym 'XX]{Make sure to enter the correct
  conference title from your rights confirmation emai}{June 03--05,
  2018}{Woodstock, NY}
\acmPrice{15.00}
\acmISBN{978-1-4503-XXXX-X/18/06}

\begin{document}
\newsavebox\CBox
\def\textBF#1{\sbox\CBox{#1}\resizebox{\wd\CBox}{\ht\CBox}{\textbf{#1}}}

\title{You Do Not Need Additional Priors in Camouflage Object Detection}

% \author{Anonymous authors}
%  \affiliation{
%    \institution{Paper under double-blind review}
%  }
% \renewcommand{\shortauthors}{Anonymous Author, et al.}

% \author{Anonymous authors}
%  \affiliation{
%    \institution{Paper under double-blind review}
%  }
% \renewcommand{\shortauthors}{Anonymous Author, et al.}

% \author{Anonymous authors}
%  \affiliation{
%    \institution{Paper under double-blind review}
%  }
% \renewcommand{\shortauthors}{Anonymous Author, et al.}

% \author{Anonymous authors}
%  \affiliation{
%    \institution{Paper under double-blind review}
%  }
% \renewcommand{\shortauthors}{Anonymous Author, et al.}

% \author{Anonymous authors}
%  \affiliation{
%    \institution{Paper under double-blind review}
%  }
% \renewcommand{\shortauthors}{Anonymous Author, et al.}

% \author{Anonymous authors}
%  \affiliation{
%    \institution{Paper under double-blind review}
%  }
% \renewcommand{\shortauthors}{Anonymous Author, et al.}

\author{Yuchen Dong}
\email{dongyc91@outlook.com}
\orcid{0000-0001-8155-5267}
\affiliation{%
  \institution{Academy of Military Sciences}
  \city{Beijing}
  \country{China}
  \postcode{100000}
}

\author{Heng Zhou}
\affiliation{%
  \institution{School of Electronic Engineering, Xidian University}
  \city{Xi'an}
  \country{China}}
\email{hengzhou@stu.xidian.edu.cn}

\author{Chengyang Li}
\affiliation{%
  \institution{School of Computer Science, Peking University}
  \city{Beijing}
  \country{China}}
\email{chengyang_li@stu.pku.edu.cn}

\author{Junjie Xie}
\affiliation{%
  \institution{Academy of Military Sciences}
  \city{Beijing}
  \country{China}
}
\email{xiejunjie06@gmail.com}

\author{Yongqiang Xie}
\affiliation{%
 \institution{Academy of Military Sciences}
 \city{Beijing}
 \country{China}}
 \email{yqxie.ams@gmail.com}

\author{Zhongbo Li}
\affiliation{%
  \institution{Academy of Military Sciences}
  \city{Beijing}
  \country{China}}
  \email{zbli83@foxmail.com}


\begin{abstract}
Camouflage object detection (COD) poses a significant challenge due to the high resemblance between camouflaged objects and their surroundings. Although current deep learning methods have made significant progress in detecting camouflaged objects, many of them heavily rely on additional prior information. However, acquiring such additional prior information is both expensive and impractical in real-world scenarios. Therefore, there is a need to develop a network for camouflage object detection that does not depend on additional priors.
In this paper, we propose a novel adaptive feature aggregation method that effectively combines multi-layer feature information to generate guidance information. In contrast to previous approaches that rely on edge or ranking priors, our method directly leverages information extracted from image features to guide model training. Through extensive experimental results, we demonstrate that our proposed method achieves comparable or superior performance when compared to state-of-the-art approaches.
\end{abstract}




% \centering
%   \includegraphics[width=0.8\textwidth,height=0.5\textwidth]{paper/picture/additional_prior.pdf}
%   \caption{From left to right, two common types of additional prior information are listed, namely the boundary prior and the ranking prior.}
%   \label{prior}

\received{20 February 2007}
\received[revised]{12 March 2009}
\received[accepted]{5 June 2009}

\maketitle
\begin{figure}[htbp]
	\centerline{\includegraphics[width=\columnwidth]{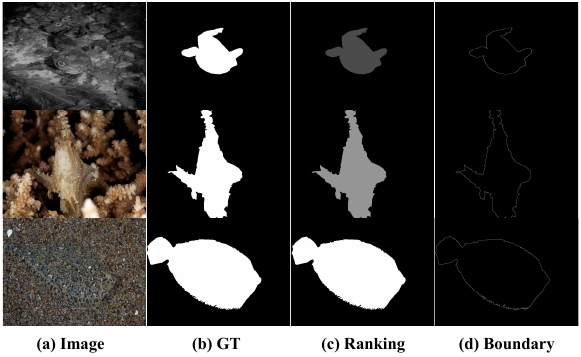}}
	\caption{We introduce two common types of additional prior information: boundary priors and ranking priors. Specifically, the boundary prior refers to boundary information about the concealed object, while the ranking prior includes information predefined based on the overall difficulty of human observation of the concealed object.}
	\label{prior}
\end{figure}

\section{Introduction}
Camouflaged object detection is a research field that focuses on the identification of camouflaged objects within images \cite{yang2021uncertainty, he2023camouflaged}. Specifically, camouflaged objects can be categorized into natural camouflage and artificial camouflage \cite{fan2021concealed}. Natural camouflage refers to how organisms utilize properties such as brightness, color, and body shape to conceal themselves in their environment \cite{troscianko2009camouflage}. Many species employ this strategy as a means of survival, either to avoid predators or to capture prey \cite{cuthill2005disruptive, zhang2022preynet}. On the other hand, artificial camouflage involves the use of techniques such as paint, clothing, and accessories to create visual deception and conceal the object. This form of camouflage is commonly employed in military applications \cite{zheng2018detection, fang2019camouflage}, art \cite{chu2010camouflage}, and other domains \cite{xie2020segmenting, fan2020pranet}.
The primary objective of camouflaged object detection is to generate a binary detection map for the camouflaged objects present in the image \cite{fan2021concealed}. However, detecting camouflage presents a challenge due to the visual similarity between the foreground and background \cite{liu2022modeling, ji2022fast, lv2023towards}.

Some recent studies \cite{lv2021simultaneously,kajiura2021improving} have tackled the challenge of detecting camouflaged objects by incorporating additional priors. These researchers argue that relying solely on image features for training is insufficient to accurately identify boundary information and comprehend camouflaged objects \cite{lv2021simultaneously}. Fig. \ref{prior} illustrates common additional priors, including boundary and rank information. Specifically, \cite{kajiura2021improving} utilized boundary information to enhance the model's ability to detect object edges, while \cite{lv2021simultaneously} employed rank information to assist the model in understanding the evolution of camouflaged objects and animals. However, it is crucial to note that manual annotation is required for these additional priors, which increases the implementation cost of utilizing such models.

To tackle this challenge, we introduce a novel adaptive feature fusion module (AFFM) that generates guidance information. Our objective is to utilize the guidance information derived from image features to guide the model's learning process, eliminating the need for additional priors. The AFFM is designed to effectively combine and integrate features from different network layers, enabling the model to exploit the inherent relationships among these features. Additionally, we propose a \textbf{P}rogressive \textbf{F}eature \textbf{R}efinement \textbf{Net}work (PFRNet) that utilizes the guidance information to direct the model in the progressive refinement of image features. Through extensive experiments, we provide compelling evidence of the effectiveness and robustness of our proposed approach.

Overall, our work makes the following main contributions:

$\bullet$ Firstly, we introduce an adaptive feature fusion module that extracts guidance information directly from image features, eliminating the need for additional priors. 

$\bullet$ Secondly, we propose a novel PFRNet that enhances the performance of camouflaged object detection by leveraging guidance information to refine the image features.

$\bullet$ Thirdly, we introduce two modules: the Feature Refinement Module (FRM) and the Context-Aware Feature Decoding Module (CFDM). These modules utilize the guidance information to refine features, guide model training, and efficiently integrate the refined features using a cross-channel approach, resulting in precise predictions.

\section{Related Work}
\textbf{Camouflage Object Detection.} 
In recent years, COD methods can be categorized into two groups: those utilizing prior information and those not relying on prior information.

\textbf{With addition prior.}
\textit{Lv et al.} \cite{lv2021simultaneously}  introduced a joint framework for locating, segmenting, and ranking camouflaged objects. They incorporated additional ranking information during training, which improved the understanding of camouflage. 
\textit{Zhai et al.} \cite{zhai2021mutual} proposed mutual graph learning, which effectively separates images into task-specific feature maps for precise localization and boundary refinement. 
\textit{Sun et al.} \cite{sun2022boundary} explored the use of object-related edge semantics as an additional guide for model learning, encouraging the generation of features that emphasize the object's edges.
\textit{He et al.} \cite{he2022eldnet} suggested using edge likelihood maps for guiding the fusion of camouflaged object features, aiming to enhance detection performance by improving boundary details.
\textit{Kajiura et al.} \cite{kajiura2021improving} employed a pseudo-edge generator to predict edge labels, contributing to accurate edge predictions.
\textit{Zhu et al.} \cite{zhu2021inferring} proposed the utilization of Canny edge \cite{canny1986computational} and ConEdge techniques to assist in model training.
\textit{Li et al.} \cite{li2021uncertainty} proposed co-training camouflage and saliency objects \cite{feng2019attentive, perazzi2012saliency} to enhance model detection.
\textit{Yang er al.} \cite{yang2021uncertainty} integrated the advantages of Bayesian learning and transformer-based \cite{han2021transformer, han2022survey} inference.They introduced Uncertainty-Guided Random Masking as prior knowledge to facilitate model training.
However, prior information is often expensive and impractical.

\textbf{Without additional prior.}
%PFNet
\textit{Mei et al.} \cite{mei2021camouflaged} proposed a localization module, a focus module, and a novel distraction mining strategy to enhance model performance. 
%SINet
\textit{Fan et al.} \cite{fan2020camouflaged, fan2021concealed} introduced a search and recognition network inspired by the predatory behavior of hunters in nature, which implements object localization and recognition steps.
% However, only some algorithms get satisfactory results by simply augmenting features with a feature stack. 
% c2FNet
\textit{Sun et al.} \cite{sun2021context} proposed an attention-inducing fusion module that integrates multi-level features and incorporates contextual information for more effective prediction.
\textit{Zhang et al.} \cite{zhang2022preynet} proposed a model that incorporates two processes of predation, specifically sensory and cognitive mechanisms. To achieve this, specialized modules were designed to selectively and attentively aggregate initial features using an attention-based approach. 
\textit{Jia et al.} \cite{jia2022segment} proposed a method where the model attends to fixation and edge regions and utilizes an attention-based sampler to progressively zoom in on the object region instead of increasing the image size. This approach allows for the iterative refinement of features.
In practice, algorithms that do not depend on prior information typically utilize various techniques to aggregate features with different receptive field sizes to obtain better detection results. While these algorithms are cost-effective, they often face limitations in efficiently mining image information. In contrast, our proposed PFRNet generates guidance information by extracting valuable features from the image itself without relying on prior information, thereby effectively guiding the model training process.

\begin{figure*}[!htbp]
        \centering
	\includegraphics[width=\linewidth]{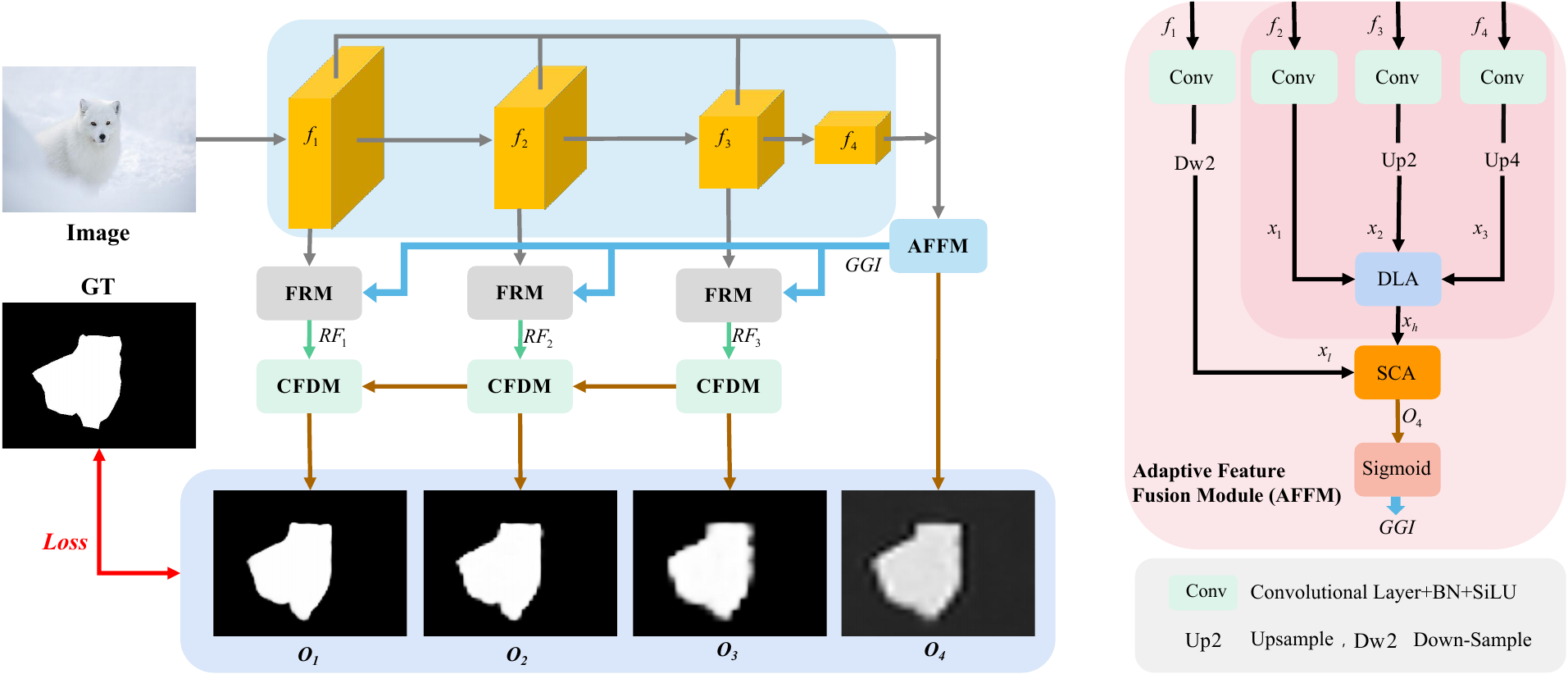}
	\caption{\textbf{Overview of our framework.}
				The proposed PFRNet framework comprises three novel components: the adaptive feature fusion module (AFFM), the feature refinement module (FRM), and the context-aware feature decoding module (CFDM).  
    The AFFM is responsible for learning the correlations between multi-layer features and adaptively fusing them to generate reliable guidance information. This guidance information is robust enough to replace the need for additional prior information during model training. The FRM utilizes the guidance information to refine image features, while the CFDM considers the relationships between different branches of the same feature and decodes features top-down, incorporating contextual semantics across channels.
    %                 The AFFM is responsible for learning the correlations between multi-layer features and adaptively fusing them to generate reliable guidance information. This guidance information is robust enough to replace the need for additional prior information during model training.
				% The FRM utilizes the guidance information to refine image features.
				% The CFDM considers the relationships between different branches of the same feature and decodes features top-down with a combination of contextual semantics across channels.
    }
	\label{fig_total}
\end{figure*}

\section{Proposed Method}

\subsection{Overall Architecture}

The architecture of PFRNet is illustrated in Fig.~\ref{fig_total}, which consists of three modules:  adaptive feature fusion module (AFFM), feature refinement module (FRM), and context-aware feature decoding module (CFDM), as described in Sec.~\ref{AFFM}, Sec.~\ref{FRM} and Sec.~\ref{CFDM} respectively.
For extracting multi-scale features, the Res2Net-50 \cite{gao2019res2net} architecture is utilized as the backbone. In this paper, the multi-scale features are obtained from the last four layers of the feature hierarchy. The layer closest to the input is excluded as it contains excessive noise and has a small receptive field. Please note that the layer closest to the input is not depicted in Fig. \ref{fig_total}.

\subsection{Adaptive Feature Fusion Module} \label{AFFM}
The boundary prior is used to aid the object detection model in localizing and segmenting objects \cite{zhao2019egnet}. However, the necessary information for localization and segmentation is already present in the image features. Additional prior information is utilized because the image features often contain significant noise, which makes it challenging for previous models to extract reliable feature information.
In our proposed module, we employ an adaptive feature fusion approach that learns the relationships between different feature layers. This enables the effective fusion of features by considering the importance of each layer. As a result, we prioritize the most discriminative and complementary information from each layer while suppressing noisy and irrelevant features.
% $\{f_i\}_{i=2}^4 \in \mathbb{R}^{\frac{H}{2^{i+1}}\times \frac{W}{2^{i+1}}\times 256i}$
Specifically, AFFM combines high-level features $\{f_i\}_{i=2}^4$ and low-level features $f_1$ to generate effective global guidance information. This global guidance information assists the features at each layer in complementing missing information.
% This result \footnote{Details of the aggregation experiments are provided in the Appendix.} shows that aggregating all features gives a more accurate result.
% It has been shown that aggregating all the features gives more accurate results, as detailed in the appendix.

% As shown in Fig. \ref{fig_total}, AFFM first aggregates the high-level features $\{f_i\}_{i=2}^4$ via deep layer attention (DLA) \cite{niu2020single} and then fuses the low-level features $f_1$ via spatial channel attention (SCA) to generate the output $O_4$. $O_4$ gets the global guidance information via the sigmoid function.

In Fig. \ref{fig_total}, the AFFM is depicted, which consists of two branches. Firstly, applying a convolution operation to all input features. Subsequently, the higher-level features are upsampled to achieve a uniform size $\{x_{i}\}_{i=1}^3 \in \mathbb{R}^{\frac{H}{8} \times \frac{W}{8} \times 256}$. The feature maps of the same size undergo the deep layer attention (DLA) \cite{niu2020single} operation, which explores the interrelationships between different layers. The resulting features are aggregated to produce $x_h \in \mathbb{R}^{\frac{H}{8} \times \frac{W}{8} \times 768}$. 
The computation process of DLA is represented by Eq. \ref{equa1}.
% As shown in Fig. \ref{fig_total}, AFFM is divided into two branches. First, a convolution operation is done on all input features. Then, the higher-level features are upsampled to obtain a uniform size $\{x_{i}\}_{i=2}^4 \in \mathbb{R}^{\frac{H}{8} \times \frac{W}{8} \times 256}$. The deep layer attention (DLA) \cite{niu2020single} operation is performed on feature maps of the same size. The interrelationship between different layers is learned through DLA, and the features are aggregated to obtain $x_h \in \mathbb{R}^{\frac{H}{8} \times \frac{W}{8} \times 768}$. The calculation of DLA is shown in Eq. \ref{equa1}.
\begin{equation}
	\label{equa1}
	\begin{split}
		&w_{i,j} = Softmax(\phi(x)_i \cdot (\phi(x))_j^T), i,j \in\{1,2,3\}  \\
		&x_j = \beta \sum_{i=1}^3 w_{i,j} x_i + x_j ,x_i/x_j \in \{x_1,x_2,x_3\}\\
		&x_h = [x_1;x_2;x_3]
	\end{split}
\end{equation}
where $w_{i,j}$ represents the correlation weight between layer i and layer j, $\phi(\cdot)$ denotes the reshape operations, and $\beta$ is initially set to 0 and then automatically assigned by the network.

To derive feature $x_l\in \mathbb{R}^{\frac{H}{8} \times \frac{W}{8} \times 128}$ from the low-level feature $f_1$, we use operations involving $1\times 1$ convolution and downsampling. Subsequently, a spatial channel attention (SCA) operation is performed on features $x_l$ and $x_h$, resulting in the feature $O_4 \in \mathbb{R}^{\frac{H}{8} \times \frac{W}{8} \times 1}$. This $O_4$ feature is one of the output features generated by the model. The above process can be described as follows:
% For the low-level feature $f_1$, use $1\times 1$ convolution and downsampling operations to obtain $x_l\in \mathbb{R}^{\frac{H}{8} \times \frac{W}{8} \times 128}$. 
% % \begin{equation}
% % 	x_l = Dw_2(Conv_{1\times 1}(f_1))
% % \end{equation}
% % Where $Conv_{i\times i}$ represents a set of convolution operations with a convolution kernel size of $i\times i$, a BN layer, and a SiLU activation function, $ Dw_j $ represents a sampling operation with a downsampling ratio of j.
% After that, features $x_l$ and $x_h$ undergo spatial channel attention (SCA) operations to obtain $O_4 \in \mathbb{R}^{\frac{H}{8} \times \frac{W}{8} \times 1}$, which is one of the output features of this model. 
% It is noted that CBAM \cite{woo2018cbam} in SCA mainly extracts features from the spatial and channel dimensions.
\begin{equation}
    \label{form_o4}
    \begin{split}
        &x_l = Conv_{1\times 1}(f_1) \\
        &O_4 = SCA(x_l,x_h) \\
        &SCA \Leftarrow Conv_{1\times 1}(Conv_{3\times 3}(CBAM(Conv_{1\times 1}(\cdot))))
	%O_4=Conv_{1\times 1}(Conv_{3\times 3}(CBAM(Conv_{1\times 1}(x_l,x_h))))
    \end{split}    
\end{equation}
Where $Conv_{i\times i}$ represents a set of convolution operations with a convolution kernel size of $i\times i$, a BN (Batch Normalization) layer, and a SiLU activation function. 
We apply CBAM \cite{woo2018cbam} to extract features from both the spatial and channel dimensions. 
Finally, a sigmoid calculation is performed on the resulting feature $O_4$ to obtain the global guide information ($\mathit{GGI}$).

% The aforementioned process enables the model to develop a more comprehensive understanding of the image.

% Through the integration of high-level and low-level features within the AFFM, a more comprehensive understanding of the image can be achieved. The generated guidance features utilize discriminative and complementary information from different feature layers to effectively guide model training.
The integration of high and low-level features in the AFFM module enables a more comprehensive understanding of the image. By combining features from different layers, the model can leverage the unique strengths of each layer. This adaptive fusion process ensures that the generated guidance information provides valuable guidance during the model's training process, utilizing discriminative and complementary information from the various feature layers. The fusion of features enhances the model's capacity to capture relevant information, thereby improving its performance in camouflaged object detection tasks.

% As previously mentioned, integrating high-level and low-level features in AFFM facilitates a more comprehensive understanding of the image. This adaptive fusion process ensures that each layer effectively leverages its unique strengths, thereby serving as a valuable guide for model training.

% As mentioned earlier, global guidance encompasses understanding the overall image semantics and synthesising contextual information. Effective utilization of global guidance information can significantly enhance the detection capabilities of the model.
% Combining low-level features and high-level features, the guidance information contains complete edge, localization, semantic and other information. 
% \begin{equation}
% 	\label{form_ggi}
% 	GGI=Sigmoid(O_4)
% \end{equation}

\subsection{Feature Refinement Module} \label{FRM}
\begin{figure}
	\centerline{\includegraphics[width=\columnwidth]{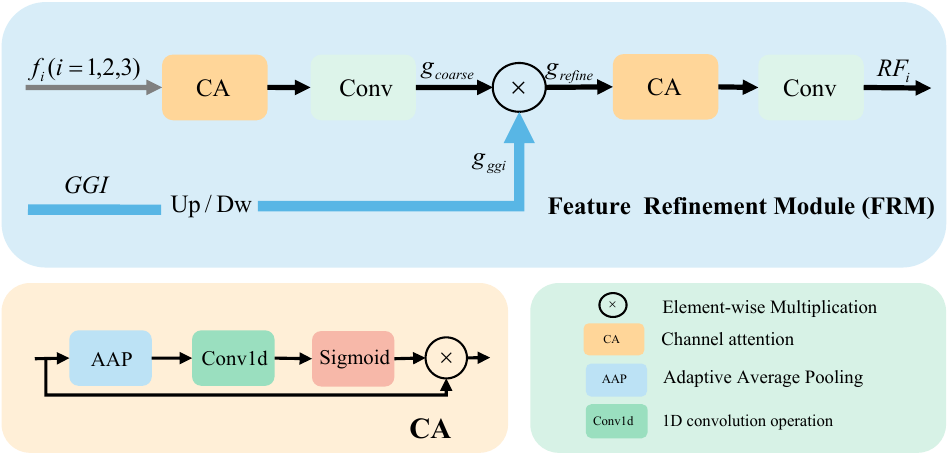}}
        \caption{Illustrate of FRM. FRM uses $\mathit{GGI}$ to refine the image features and complement the missing information in each layer of features. The guidance provided by $\mathit{GGI}$ assists the entire network in conducting a more comprehensive image content analysis.}
	\label{fig_frm}
\end{figure}

As mentioned, the AFFM module extracts relevant object details from image features to achieve a comprehensive understanding of the entire image, generating global guidance information ($\mathit{GGI}$).
In the feature refinement module (FRM), we incorporate the generated $\mathit{GGI}$ into the model to guide the training of the detection model. Specifically, we utilize $\mathit{GGI}$ to refine the image features at each layer, complementing the missing information in each layer of features. Fig. \ref{fig_frm} provides an overview of the overall structure of the FRM.

In FRM, the image features $\{f_i\}_{i=1}^3$ undergo channel attention (CA) \cite{wang2020eca} to learn cross-channel correlations, enhancing the expressiveness of the features. Subsequently, a convolution operation is applied to reduce the number of feature channels, resulting in coarse feature $g_{coarse} \in \mathbb{R}^{\frac{H}{2^{i+1}}\times \frac{W}{2^{i+1}}\times 256}$. 
Simultaneously, feature $\mathit{GGI}$ performs (up/down) sampling operations to obtain feature $g_{ggi} \in \mathbb{R}^{\frac{H}{2^{i+1}}\times \frac{W}{2^{i+1}}\times 1}$. 
The refinement features $g_{refine}$ is obtained by element-wise multiplying $g_{coarse}$ and $g_{ggi}$. 
Finally, features $\{RF_i\}_{i=1}^3$ are calculated by applying CA and convolution operations to $g_{refine}$. The calculation process is depicted in Eq. \ref{equ_cr}.
\begin{equation}
	\begin{split}
		&g_{coarse} = Conv_{3\times 3}(CA(f_i)),i \in \{1,2,3\} \\
		&g_{refine} = g_{coarse} \otimes Up_2/Dw_2(g_{ggi}) \\
		&RF_i = Conv_{1\times 1} (CA(g_{refine})) ,i \in \{1,2,3\}
	\end{split}
	\label{equ_cr}
\end{equation}
where $f_i$ represents the image features output by the backbone. $ RF_i $ is the refined feature.

\subsection{Context-aware Feature Decoding Module}\label{CFDM}
\begin{figure}
	\centerline{\includegraphics[width=\columnwidth]{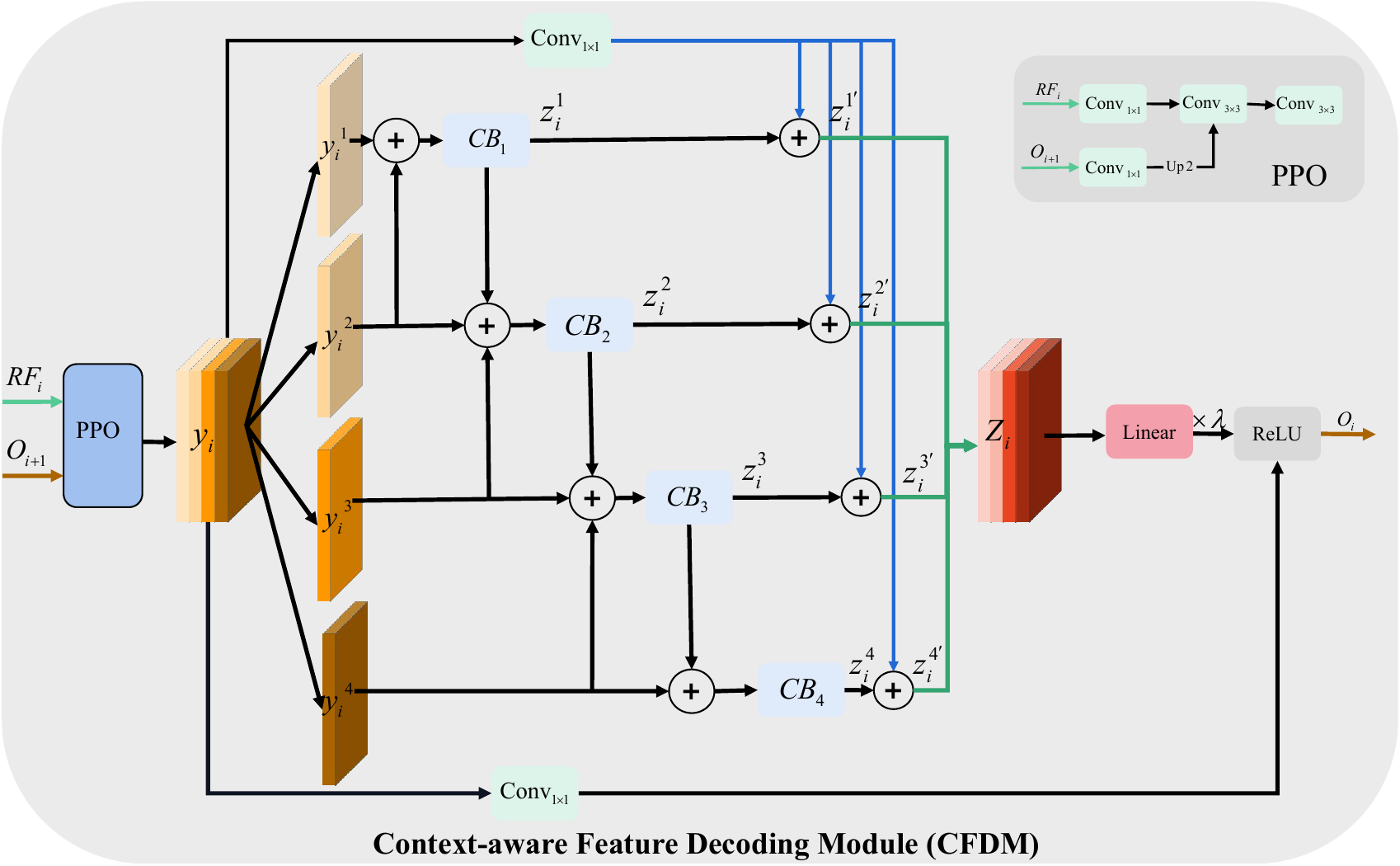}}
        \caption{Illustration of CFDM. CFDM obtains multi-scale contextual features between adjacent features through cross-channel interaction learning, enhancing the feature representation.}
	\label{fig_cdm}

\end{figure}

To integrate the contextual features and generate informative predictive binary maps, we introduce the context-aware feature decoding module (CFDM). In contrast to the Texture Enhanced Module (TEM) \cite{fan2021concealed}, our approach considers the semantic correlation between different branches within the same layer of features. 
As shown in Fig. \ref{fig_cdm}, we obtain features $\{y_i\}_{i=1}^3 \in \mathbb{R}^{\frac{H}{2^{i+1}}\times \frac{W}{2^{i+1}}\times 256}$ by aggregating the fine features $\{RF_i\}_{i=1}^3$ and the high-level output features $\{O_i\}_{i=2}^3$ through preprocessing operation (PPO). The feature $\{y_i\}_{i=1}^3$ comprises top-down semantic information and is subsequently divided equally into four parts $[y_i^1; y_i^2; y_i^3; y_i^4]$ in the channel dimension. 
The generation process of $\{y_i\}_{i=1}^3$ is illustrated in Eq. \ref{yi}.
\begin{equation}
	\label{yi}
	\begin{split}
		&y_i = PPO(RF_i,O_{i+1}),i\in \{1,2\} \\
		&y_3 = RF_3 % &PPO(x,y)  = Conv_{3\times 3}(Conv_{3\times 3}(Conv_{1\times 1}(x)\oplus Up_2(Conv_{1\times 1}(y))))
	\end{split}
\end{equation}
After obtaining the feature $y_i^j$, we add the features of a branch to the features of its neighboring branches. This process can be formulated as follows:
\begin{equation}
	\label{form_2}
	\begin{split}
		z_i^1 &= CB_1(y_i^1 + y_i^2) ,i\in\{1,2,3\} \\
		z_i^{j}&= CB_j(z_i^{j-1 } + y_i^j + y_i^{j+1}),i\in\{1,2,3\},j\in \{2,3\} \\
		z_i^4&=CB_4(z_i^3 + y_i^4 ) ,i\in\{1,2,3\}
	\end{split}
\end{equation}
where $\{CB_j\}_{j=1}^4$ indicates a series of convolution operations and the specific composition of $\{CB_j\}_{j=1}^4$ is shown in Eq. \ref{form_3}.
%In CFD, we use $\{CB_j\}_{j=1}^4$ to decode the contextual features.
\begin{equation}
	\label{form_3}
	\begin{split}
		CB_1 &\Leftarrow DConv_{3 \times 3}^{1}(Conv_{1\times 1}(\cdot)) \\
		CB_2 &\Leftarrow DConv_{3 \times 3}^{3}(Conv_{3 \times 1}(Conv_{1\times 1}(\cdot))) \\
		CB_3 &\Leftarrow DConv_{3\times 3}^{3}(Conv_{1 \times 3}(Conv_{1\times 1}(\cdot))) \\
		CB_4 &\Leftarrow DConv_{3\times 3}^{5}(Conv_{1 \times 3}(Conv_{3\times 1}(Conv_{1\times 1}(\cdot))))
	\end{split}
\end{equation}
where $DConv_{i\times i}^j$ represents a atrous convolution \cite{gu2019net} with a convolution kernel size of $i \times i$ and a dilation rate of $j$.

Subsequently, we apply residual connections on each branch to obtain the feature $z_i^{j \textquoteright} \in \mathbb{R}^{\frac{H}{2^{i+1}}\times \frac{W}{2^{i+1}}\times 256} $. Then, we merge all $z_i^{j \textquoteright}$ to generate the feature $\{Z_i\}_{i=1}^3 \in \mathbb{R}^{\frac{H}{2^{i+1}}\times \frac{W}{2^{i+1}}\times 256}$. Finally, we use a linear function, ReLU function, and a residual connection to obtain output feature $\{O_i\}_{i=1}^3 \in \mathbb{R}^{\frac{H}{2^{i+1}}\times \frac{W}{2^{i+1}}\times 1}$.
The scaling factor $\lambda$ = 0.5 is employed in this process. 

\begin{table*}[!ht]
	\centering
	\caption{Quantitative comparison with state-of-the-art methods for COD on three benchmarks using four widely used evaluation metrics (i.e.,$S_{\alpha}$, $E_{\phi}$, $F_{\beta}^w$, and $\mathcal{M}$). The best scores are highlighted in $\textBF{Bold}$, and the symbol $\uparrow$  indicates that a higher score is better.}
	\label{tab1}
         \setlength{\tabcolsep}{8mm}{
		\begin{tabular}{lccc}
			\toprule
                \multicolumn{1}{c}{Method}  & \multicolumn{1}{c}{CAMO-Test} & \multicolumn{1}{c}{COD10K-Test} & \multicolumn{1}{c}{NC4K} \\
			\cline{2-4} 
			\multicolumn{1}{c}{} &  $S_{\alpha} \uparrow$  $E_{\phi} \uparrow$  $F_{\beta}^w \uparrow$  $\mathcal{M} \downarrow$ & $S_{\alpha} \uparrow$  $E_{\phi} \uparrow$  $F_{\beta}^w \uparrow$  $\mathcal{M} \downarrow$ & $S_{\alpha} \uparrow$  $E_{\phi} \uparrow$  $F_{\beta}^w \uparrow$  $\mathcal{M} \downarrow$ \\
                \midrule
               2019 EGNet \cite{zhao2019egnet}  & 0.732 0.800 0.604 0.109 & 0.736 0.810 0.517 0.061 & 0.777 0.841 0.639 0.075 \\
			2019 SCRN \cite{wu2019stacked} & 0.779 0.797 0.643 0.090 & 0.789 0.817 0.575 0.047& 0.830 0.854 0.698 0.059 \\
			2020 $\mathrm{F^3 NET}$ \cite{wei2020f3net} & 0.711 0.741 0.564 0.109 & 0.739 0.795 0.544 0.051 & 0.780 0.824 0.656 0.070 \\
			2020 CSNET \cite{gao2020highly} & 0.771 0.795 0.642 0.092 & 0.778 0.809 0.569 0.047 & 0.750 0.773 0.603 0.088 \\
			2020 BASNET \cite{qin2021boundary} & 0.749 0.796 0.646 0.096 & 0.802 0.855 0.677 0.038 & 0.817 0.859 0.732 0.058 \\
			2020 SINet \cite{fan2020camouflaged} & 0.745 0.804 0.644 0.092 & 0.776 0.864 0.631 0.043 & 0.808 0.871 0.723 0.058 \\
			2021 PFNet \cite{mei2021camouflaged} & 0.782 0.841 0.695 0.085 & 0.800 0.877 0.660 0.040 & 0.829 0.887 0.745 0.053 \\
			2021 S-MGL \cite{zhai2021mutual} & 0.772 0.806 0.664 0.089 & 0.811 0.844 0.654 0.037 & 0.829 0.862 0.731 0.055 \\
			2021 $\mathrm{C^2FNet}$ \cite{sun2021context} & 0.799 0.851 0.710 0.078 & 0.811 0.886 0.669 0.038 & 0.843 0.899 0.757 0.050\\
			2022 BGNet \cite{sun2022boundary} & 0.807 0.861 0.742 0.072 & 0.829 \textBF{0.898} \textBF{0.719} 0.033 & 0.849 \textBF{0.903} 0.785 0.045 \\
			\midrule
			PFRNet (Ours) & \textBF{0.827} \textBF{0.877} \textBF{0.754} \textBF{0.069} & \textBF{0.833} 0.888 0.710 \textBF{0.033} & \textBF{0.859} 0.902 \textBF{0.785} \textBF{0.045}  \\
			  \bottomrule
		\end{tabular}}
\end{table*}

\subsection{Loss Function}
PFRNet incorporates two types of loss functions: dice loss ($L_{dice}$) \cite{xie2020segmenting} and structural loss ($L_{struct}$) \cite{wei2020f3net}. For $O_4$, we utilize $L_{dice}$ to balance scenarios where positive and negative samples are unbalanced. For $\{O_i\}_{i=1}^3$, we apply $L_{struct}$ to promote structural consistency and accuracy.
% , as expressed in Eq. \ref{form_6}. 
\begin{equation}
	\label{form_6}
	L_{struct} = L_{BCE}^w + L_{IoU}^w 
\end{equation}

Therefore, the total loss is defined as in Eq. \ref{form_5}.
\begin{equation}
	\label{form_5}
	L_{total} = \sum_{i=1}^{3} L_{struct} (O_i,GT) + L_{dice} (O_4,GT)
\end{equation}
where $\{O_i\}_{i=1}^4$ represents the feature map generated by PFRNet, and GT refers to the ground truth. During the testing process, $O_1$ is used as the prediction result of the model.

\section{Experiments and Analysis}
\subsection{Experiment Setup}
We implement our model using PyTorch and utilize the Adam optimization algorithm \cite{kingma2014adam} to optimize the overall parameters. The learning rate starts at $1e^{-4}$, dividing by 10 every 50 epochs. The model is accelerated using an NVIDIA 3090Ti GPU. During the training stage, the batch size is set to 36, and the whole training takes approximately 100 epochs.

\begin{table*}[!ht]
	\caption{ Quantitative evaluation for ablation studies on three datasets. The best results are highlighted in $\textBF{Bold}$.} 	
	\label{tab2}
        \setlength{\tabcolsep}{4.5mm}{
			\begin{tabular}{clccc}
				\toprule			
				\multicolumn{1}{c}{No.}  & \multicolumn{1}{l}{Method}  & \multicolumn{1}{c}{CAMO-Test} & \multicolumn{1}{c}{COD10K-Test} & \multicolumn{1}{c}{NC4K} \\
				\cline{3-5} 
				& & $S_{\alpha} \uparrow$  $E_{\phi} \uparrow$  $F_{\beta}^w \uparrow$  $\mathcal{M} \downarrow$ & $S_{\alpha} \uparrow$  $E_{\phi} \uparrow$  $F_{\beta}^w \uparrow$  $\mathcal{M} \downarrow$ & $S_{\alpha} \uparrow$  $E_{\phi} \uparrow$  $F_{\beta}^w \uparrow$  $\mathcal{M} \downarrow$ \\
				\midrule
				\textBF{A} & Base & 0.799 0.860 0.717 0.078 & 0.801 0.867 0.655 0.038 & 0.834 0.887 0.745 0.050 \\
				\textBF{B} & Base+CFDM & 0.814 0.852 0.727 0.078 & 0.827 0.875 0.690 0.034 & 0.854 0.892 0.771 0.046 \\
				\textBF{C} & Base+AFFM+FRM & 0.818 0.870 0.746 0.072 & 0.832 0.885 0.710 \textBF{0.032} & 0.852 \textBF{0.903} 0.782 0.045 \\
				\textBF{D} & Base+FRM+CFDM & 0.822 0.864 0.745 0.074 & 0.827 0.871 0.694 0.035 & 0.855 0.892 0.773 0.047  \\
				\textBF{E} (ours) & Base+AFFM+FRM+CFDM &  \textBF{0.827} \textBF{0.877} \textBF{0.754} \textBF{0.069} & \textBF{0.833} \textBF{0.888} \textBF{0.710} 0.033 & \textBF{0.859 } 0.902 \textBF{0.785} \textBF{0.045} \\
				\bottomrule
		\end{tabular}}
\end{table*}

\subsection{Comparison with State-of-the-arts}
We evaluate our method on three benchmark datasets: CAMO \cite{le2019anabranch}, COD10K \cite{fan2020camouflaged}, and NC4K \cite{lv2021simultaneously}.
And we use four widely used and standard metrics to evaluate our method: MAE ($\mathcal{M}$) \cite{perazzi2012saliency}, weighted F-measure ($F_{\beta}^w$) \cite{margolin2014evaluate}, average E-measure ($E_{\phi}$) \cite{fan2018enhanced} and S-measure ($S_{\alpha}$) \cite{fan2017structure}.

To demonstrate the effectiveness of our proposed PFRNet, we conducted a comparative analysis by comparing its prediction results with those of ten state-of-the-art methods. The selected methods for comparison are EGNet\cite{zhao2019egnet}, SCRN \cite{wu2019stacked}, $\mathrm{F^3Net}$ \cite{wei2020f3net}, CSNet \cite{gao2020highly}, BASNet \cite{qin2021boundary}, SINet \cite{fan2020camouflaged}, PFNet \cite{mei2021camouflaged}, S-MGL \cite{zhai2021mutual}, BGNet \cite{sun2022boundary} and $\mathrm{C^2FNet}$ \cite{sun2021context}. 
For a fair comparison, the prediction results of these methods were provided by the original authors or generated using models trained with open-source code.\\

\textbf{Quantiative Evaluation}
Table \ref{tab1} summarizes the quantitative results of different COD methods on the three benchmark datasets. Our method outperforms previous methods in all four evaluation indicators on the three datasets. In particular, compared with the BGNet \cite{sun2022boundary}, our method shows an average increase of $1.13\% $ in $S_{\alpha}$, $0.16\% $ in $E_{\phi}$, and $0.1\%$ in $F_{\beta}^w$.

\begin{figure*}[!htbp]
    \centering
	\includegraphics[width=\linewidth]{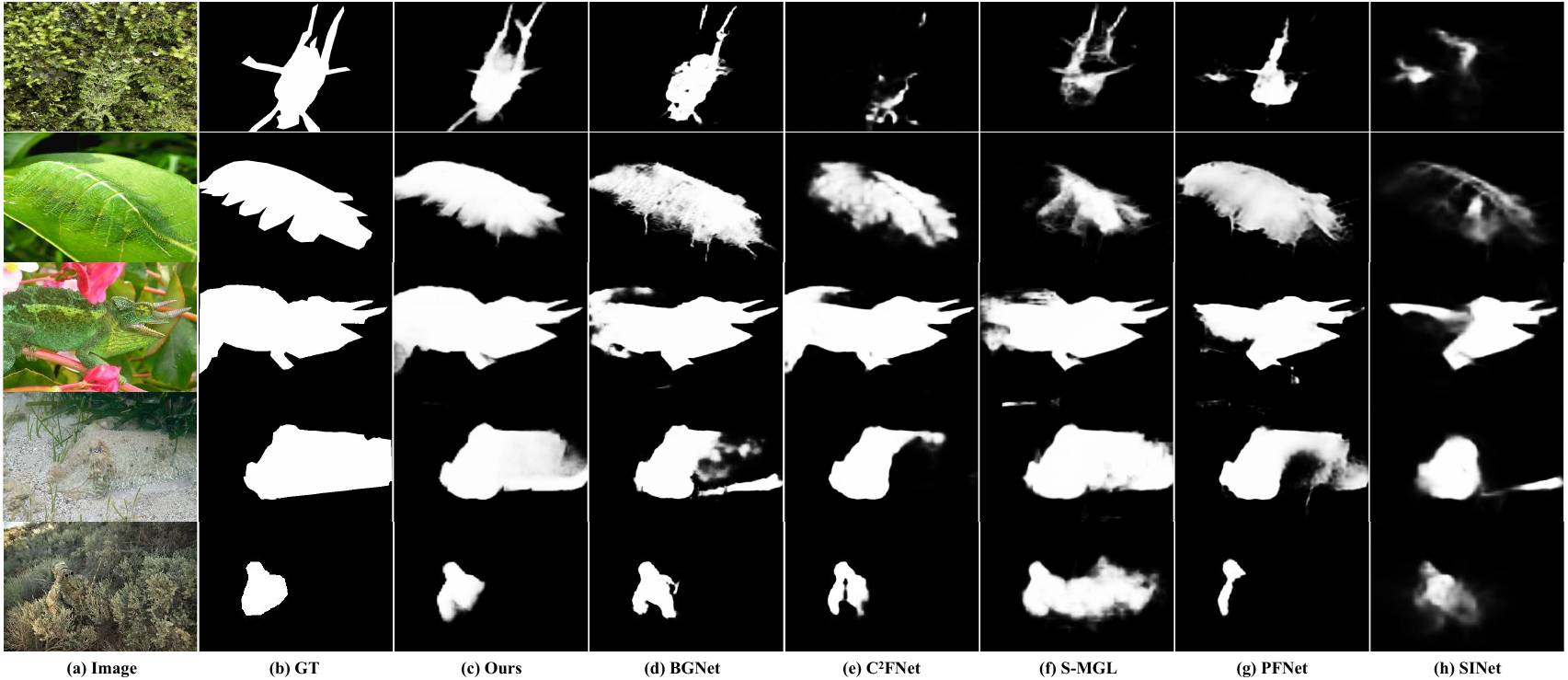}
	\caption{Visual comparison of the proposed model with five state-of-the-art COD methods.}
	\label{fig_cop}
\end{figure*}

Regarding the slightly lower training results compared to BGNet for some datasets, we have  analyzed the reasons behind these differences. We attribute them to two main factors.
Firstly, during supervised learning, we did not incorporate additional priors, which may have influenced the detection performance of our model. The absence of these additional priors might have affected the model's ability to capture certain intricate details and characteristics of the objects.
Secondly, as illustrated in Fig. \ref{fig_cop}, when comparing the prediction results of PFRNet with those of other models, we observed that PFRNet effectively addressed the issue of uneven pixel distribution. However, this occasionally led to an expanded range of predicted objects, which might have impacted specific evaluation metrics. For instance, in the first row of Fig. \ref{fig_cop}, the leg of the animal and in the fifth row, the size of the people were slightly affected. \\
\textbf{Qualitative Evaluation}
Fig. \ref{fig_cop} shows a qualitative comparison of different COD methods on the CAMO dataset. 
Compared to several other state-of-the-art models, our proposed PFRNet effectively addresses the issue of uneven image distribution. This can be observed in the fourth and fifth rows of Fig. \ref{fig_cop}, where PFRNet's predictions exhibit improved completeness and accuracy regarding boundary processing, leading to visually superior results. 
Furthermore, this comparison highlights the effectiveness of our adaptive feature fusion approach in extracting valuable edge information. Our model achieves superior edge detection performance without requiring additional edge prior guidance.

\begin{table*}[!ht]
	\centering
	\caption{\textBF{Sensitivity analysis on $\lambda$.} We compared the best scale factors using the four widely used indicators on three datasets. The best results are highlighted in $\textBF{Bold}$.}
	\label{tab4}
        \setlength{\tabcolsep}{8mm}{
		\begin{tabular}{ccccc}
                \toprule
                \multicolumn{1}{c}{No.}  & \multicolumn{1}{c}{$\lambda$} & \multicolumn{1}{c}{CAMO-Test} & \multicolumn{1}{c}{COD10K-Test} & \multicolumn{1}{c}{NC4K} \\
			\cline{3-5}  
			& & $S_{\alpha} \uparrow$  $E_{\phi} \uparrow$  $F_{\beta}^w \uparrow$  $\mathcal{M} \downarrow$ & $S_{\alpha} \uparrow$  $E_{\phi} \uparrow$  $F_{\beta}^w \uparrow$  $\mathcal{M} \downarrow$ & $S_{\alpha} \uparrow$  $E_{\phi} \uparrow$  $F_{\beta}^w \uparrow$  $\mathcal{M} \downarrow$ \\
			\midrule
			% 1 & 0.1 & 0.720 0.772 0.573 0.127 & 0.724 0.773 0.504 0.065 & 0.739 0.789 0.575 0.096 \\
			\uppercase\expandafter{\romannumeral1} & 0.2 & 0.817 0.863 0.740 0.075 & 0.830 0.877 0.705 0.033 & 0.855 0.898 0.781 0.045 \\
			\uppercase\expandafter{\romannumeral2} & 0.3 & 0.826 0.875 0.753 0.070 & 0.832 0.884 0.703 0.032 & 0.857 0.899 0.785 \textBF{0.044} \\
			\uppercase\expandafter{\romannumeral3} & 0.4 & 0.825 0.869 0.750 0.070 & 0.832 0.885 0.708 \textBF{0.032} & 0.855 0.897 0.782 0.045 \\
			\uppercase\expandafter{\romannumeral4} & 0.5 & \textBF{0.827} \textBF{0.877} \textBF{0.754} \textBF{0.069} & \textBF{0.833} \textBF{0.888} \textBF{0.710} 0.033 & \textBF{0.859} \textBF{0.902} \textBF{0.785} 0.045 \\
			\uppercase\expandafter{\romannumeral5} & 0.6 & 0.815 0.865 0.738 0.076 & 0.832 0.888 0.706 0.033 & 0.857 0.901 0.784 0.045 \\
			\bottomrule 
	\end{tabular}}
\end{table*}

\begin{table*}[!ht]
	\caption{ Quantitative evaluation for ablation studies on three datasets. The best results are highlighted in $\textBF{Bold}$.} 	
	\label{tab2}
        \setlength{\tabcolsep}{4.5mm}{
			\begin{tabular}{clccc}
				\toprule			
				\multicolumn{1}{c}{No.}  & \multicolumn{1}{l}{Method}  & \multicolumn{1}{c}{CAMO-Test} & \multicolumn{1}{c}{COD10K-Test} & \multicolumn{1}{c}{NC4K} \\
				\cline{3-5} 
				& & $S_{\alpha} \uparrow$  $E_{\phi} \uparrow$  $F_{\beta}^w \uparrow$  $\mathcal{M} \downarrow$ & $S_{\alpha} \uparrow$  $E_{\phi} \uparrow$  $F_{\beta}^w \uparrow$  $\mathcal{M} \downarrow$ & $S_{\alpha} \uparrow$  $E_{\phi} \uparrow$  $F_{\beta}^w \uparrow$  $\mathcal{M} \downarrow$ \\
				\midrule
				\textBF{A} & Base & 0.799 0.860 0.717 0.078 & 0.801 0.867 0.655 0.038 & 0.834 0.887 0.745 0.050 \\
				\textBF{B} & Base+CFDM & 0.814 0.852 0.727 0.078 & 0.827 0.875 0.690 0.034 & 0.854 0.892 0.771 0.046 \\
				\textBF{C} & Base+AFFM+FRM & 0.818 0.870 0.746 0.072 & 0.832 0.885 0.710 \textBF{0.032} & 0.852 \textBF{0.903} 0.782 0.045 \\
				\textBF{D} & Base+FRM+CFDM & 0.822 0.864 0.745 0.074 & 0.827 0.871 0.694 0.035 & 0.855 0.892 0.773 0.047  \\
				\textBF{E} (ours) & Base+AFFM+FRM+CFDM &  \textBF{0.827} \textBF{0.877} \textBF{0.754} \textBF{0.069} & \textBF{0.833} \textBF{0.888} \textBF{0.710} 0.033 & \textBF{0.859 } 0.902 \textBF{0.785} \textBF{0.045} \\
				\bottomrule
		\end{tabular}}
\end{table*}

\section{Conclusion}

In this paper, we propose a novel network called PFRNet for camouflaged object detection, with the goal of reducing the reliance on additional priors. Our proposed method introduces an adaptive feature fusion module that generates guidance information, eliminating the need for additional priors to guide model training. 
Additionally, we introduce two modules, FRM and CFDM, which leverage guidance information to complement missing features in image representations and enhance feature representation by incorporating contextual information. Through extensive evaluations on three benchmark datasets, we demonstrate that our approach achieves state-of-the-art performance.

% \input{paper/picture/prior}
% \input{paper/intro/introduction}
% \input{paper/relate_work/relate}
% \input{paper/picture/total_pic}
% \input{paper/second/propose}
% \input{paper/second/GGIM}
% \input{paper/second/FRM}
% \input{paper/picture/frm}
% \input{paper/second/CFDM}
% \input{paper/picture/cfdm}
% \input{paper/table/compare_result}
% \input{paper/loss/loss_func}
% \input{paper/experiment/setup}
% \input{paper/table/ablation}
% \input{paper/experiment/compare_sota}
% \input{paper/picture/contract}
% \input{paper/table/lambda}
% \input{paper/experiment/ablation}
%\input{paper/table/input_GGIM}
% \input{paper/conclusion/conclu}

\bibliographystyle{unsrt}
\clearpage

\bibliographystyle{ACM-Reference-Format}
\bibliography{file}

% \appendix









\end{document}